\documentstyle{nemlap}

\begin{document} 
\title{Stylistic Variation in an Information Retrieval Experiment}
\author{Jussi Karlgren}
\institute{Courant Institute of Mathematical Sciences \\ 
New York University}
\date{}

\maketitle

\begin{abstract}
Texts exhibit considerable stylistic variation. This paper reports an
experiment where a corpus of documents (N= 75 000) is analyzed using
various simple stylistic metrics. A subset (n = 1000) of the corpus has
been previously assessed to be relevant for answering given information
retrieval queries. The experiment shows that this subset differs
significantly from the rest of the corpus in terms of the stylistic metrics
studied. 

\end{abstract}

\section{Introduction}

Texts vary not only by topic. {\em Stylistic} variation between texts of
the same topic is often at least as noticeable as the {\em topical}
variation between texts of different topic but same genre or variety; style
is, broadly defined, the difference between two ways of saying the same
thing.

Stylistic variation in a given text, given the liberal definition above,
can occur in many ways and on many linguistic levels: lexical choice,
choice of syntactic structures, choice of cohesion markers on a textual
level, and so forth. Some choices are constrained by the intended audience
and discourse ecology the text is produced in; some are left to be entirely
determined by the author's preferences and personal idiosyncracies.  In
this experiment the stylistic variation investigated is taken from a fairly
well-edited body of text -- the Wall Street Journal -- where presumably
most writers conform to the expected norms of writing, and if not, the
texts will be edited by professional editors to conform to them. The focus
will be on stylistic variation based on the specific {\em genres} or {\em
functional styles} that occur in a daily newspaper -- as opposed to {\em
individual styles} \cite{Vachek}. Especially, the experiment investigates how the
variation relates to the usefulness of texts in a large scale information
retrieval experiment.

This experiment makes some simple measurements of {\em style markers},
indicative of stylistic variation and of genre, on a largish corpus of
documents, and compares a subset of documents that have previously been
judged relevant for answering queries in an information retrieval
experiment with the rest of the corpus.

\section{Corpus and Statistical Measurements}

The Text REtrieval Conference (TREC), organized in the form of a
competition by US government research agencies, gives participating
research organizations access to a large corpus of texts and a set of
queries that are to be used for retrieving texts from the corpus. The texts
that the organizations and systems participating in the competition suggest
as most relevant for a query are read by a number of judges, and assessed
as either relevant or not relevant. These assessments -- the {\em relevance
judgements} -- are used in this experiment to categorize the entire
corpus. Given a query, the corpus has three types of document:
relevant texts, not relevant texts, and not judged (i.e. not retrieved)
texts~\cite{TREC4}. In this experiment, if a text was judged relevant for
any query at all in the test set we will consider the text relevant.

For this experiment, a part of the TREC test corpus was selected: 74~516
Wall Street Journal articles from years the 1990, 1991, and 1992. Of these
documents, 1116 were marked relevant for at least one of fifty queries
(TREC queries 201-250) and 12~482 marked non-relevant - judged, but not
relevant for any of the queries\footnote{3493 articles were retrieved for
more than one query. Three articles were retrieved for more than 25 of 50
queries.}. Initially, the documents were analyzed to obtain simple sentence
statistics and to obtain simple measures for
syntactic complexity.

\section{Hypotheses}

The hypotheses of the experiment were 1) that certain genres or types of
text would be more likely to provide the answers the human judges would
prefer, and 2) that this preference is clear enough to be detectable even
using the fairly simple mechanisms tested in this experiment.

The stylistic variation was expected from two reasons. Firstly, by the
likelihood that the corpus contains materials that will never be useful in
a generally framed information retrieval task such as TREC: stock report
tables and the like; secondly, by the fact that the human judges, while
well trained for the task, are likely to exhibit biases for certain types
of documents, namely those which are easy to judge as being relevant or
not.

\section{Results}

The results are positive. Texts that were found relevant did differ
systematically from texts which are not found relevant; for most metrics
tested, the difference was statistically significant even by univariate
tests.\footnote{Since normal distribution assumptions cannot be expected to
hold for language data of this sort, we used the Mann Whitney U rank sum
test, which makes fewer assumptions about the values and distributions of
the variables, to test for the significance of detected differences. There
are no standard tests for multivariate significance for ``nonparametric''
variables, i.e. variables without an approximation of their value space;
this means that this experiment will miss interactions between variables,
testing them one by one, rather than risking a false positive result from
using a multivariate test based on false assumptions. The Mann Whitney U
test is one of two equivalent formulations -- the other is Wilcoxon's rank
sum test -- for calculating significant differences between some
measurement in two sets. The sets are sorted together by the result of the
measurement, and the sum of ranks is calculated for one of the
categories. If there is a difference in the measurement results between the
categories, the sum will tend to be either high or low; the thresholds for
a significant difference from the expected average value for the rank sum
are calculated using the relative sizes of the sets.} In addition, we find
that relevant texts and non-relevant texts taken together -- i.e. texts
retrieved by systems participating in the TREC evaluation -- differ from
the rest of the corpus in a systematic manner. The difference between
relevant and non-relevant texts is much smaller than the difference between
either of them and the non-judged portion of the corpus.

In summary, the results of this experiment show that retrieved highly
ranked texts -- both relevant and non-relevant -- are longer, with a more
complex sentence structure than the rest of the corpus, and that relevant
texts differ from nonrelevant in that they tend to be even more complex
textually.

\subsection{Simple statistics: Sentence Length and Word Statistics}

A simple word count reveals that relevant texts on the average are longer
than other texts -- which also has been observed, pointed out, and utilized
by the very successful Cornell research group at the latest TREC conference
\cite{Smart}. 

Word statistics -- word length, long word counts, type/token ratios -- as a
measure of terminological complexity have often been paired with sentence
length to produce readability scores and recently to study variation in
various varieties of language, as well as perform genre discrimination
\cite{Bib88,Bib89,KoC,Klare}. Table~\ref{Stats} contains a summary of the
simple statistics. Relevant texts, besides being longer, also have longer
sentences. The differences between relevant and non-relevant are
significant in a Mann Whitney test on a 95\% confidence level for all
statistics except average word length.

\begin{table}
  \begin{center}
\begin{tabular}{|l|rrrrr|} 
\hline 
Category & Number & Word count & Type-token ratio & Word length & Words per sentence \\ 
\hline \hline
Relevant    &   1116  & 755  & 0.527  & 5.08   & 19.8 \\
Misses      &  12482  & 675  & 0.551  & 5.07   & 19.3 \\
Not judged  &  60918  & 396  & 0.611  & 5.03   & 19.2 \\
\hline
\end{tabular}
  \end{center}
\caption{Simple statistics for the corpus}\label{Stats}
\end{table}

Longer texts are more likely to be relevant at least partly due to the fact
that longer texts range over several topics, and thus there is a chance
that a long text will touch a relevant topic. In this experiment, we find
that not only are relevant documents longer, but all documents retrieved by
systems, even those assessed by human judges as irrelevant, also are longer
than the average document. Not only will longer texts touch relevant topics
-- but apparently they may well touch irrelevant but confusingly similar
topics.

The non-retrieved portion of the corpus turns out to contain large numbers
of very short items, and large numbers of tables and numerical information,
both short and long, which the retrieval systems have not proffered to the
assessors for consideration. These texts presumably simple have less
topical information, and thus are hit less often by the retrieval systems
used. Running a subtopic segmentation algorithm over a number of
texts\footnote{This experiment ran on a smaller corpus of only relevant and
non-relevant documents.} produces the expected result. For the experiment
we ran a system -- TextTiles -- which cuts up a text into {\em tiles},
tentative subtopic segments \cite{Tile}. For the purposes of the
experiment, only the number of tiles were retained. The number of tiles
was higher for the relevant documents than for the non-relevant ones, but
when the experiment was rerun on texts categorized by length, we find that
long relevant texts tend to have fewer subtopics than the short ones, in
contrast with the shorter texts - see Table~\ref{Tiles}. The difference is
better than 95\% significant by Mann Whitney U for the larger number of
documents, for the long texts the risk of random results is around 10\%.

\begin{table}
  \begin{center}
\centering
\begin{tabular}{|l|rr|} 
\hline
Category     & Number & Tiles \\
\hline
\multicolumn{3}{|l|}{Documents of all lengths} \\
\hline
Relevant       & 756  & 3.8  \\ 
Not relevant   & 4406 & 3.6  \\ 
\hline
\multicolumn{3}{|l|}{Documents over a thousand words} \\
\hline
Relevant     & 176 & 8.1 \\ 
Misses       & 985 & 8.8 \\ 
\hline

\end{tabular}
\end{center}
\caption{Average number of tiles}\label{Tiles}
\end{table}

\subsection{Syntactic complexity}

Syntactic complexity is a dimension which exhibits considerable variation
between genres \cite{Losee,Mensj}. Indeed, most stylistic measures
heretofore have been attempts to find shortcuts for measuring syntactic
complexity along with lexical complexity as measured by word length and
type-token ratios. Sentence length as used above is one such method, although
arguably a blunt one -- what syntactic constructions are complex in
themselves, and when they are evidence of complexity in an already complex
subject matter is a matter of contention \cite{Daw}.

As a somewhat deeper approximation of clause complexity, we will look at the average
depth of output trees from a robust parser built for information retrieval
purposes \cite{TTP}. The parser was set to skip parsing after
a timeout threshold, and when it does so, it notes it has done so in the
parse tree. These skip marks were counted -- again, as an indication of
clausal complexity. We find as shown, in Table~\ref{Trees}, a clear
distinction between the various categories of document. Relevant documents
have, on average, deeper parse trees and more skips. Both measures show a
significant difference between relevant and non-relevant documents, again
by a Mann Whitney test on better than a 95\% confidence level.

\begin{table}
  \begin{center}
\begin{tabular}{|l|rr|} 
\hline 
Category & depth & skips \\
\hline \hline 
Relevant     &   10.0 & 0.499 \\
Non-relevant &   9.88 & 0.456  \\
Not assessed &  9.56 & 0.409 \\
\hline
\end{tabular}
\end{center}
\caption{Trees and Skips}\label{Trees}
\end{table}

\begin{table}[htbp]
  \begin{center}
\begin{tabular}{|rrl|rrrrrrr|} 
\hline 
N & Genre & Category & Tree  & Skips & Words & Tokens & Chars & Digits  & Words \\
       & &          & Depth &       &       & /Types & /Word & /kChars  & /Sent \\
\hline \hline
11 331 &  A  & & & & & & & & \\
	 330 &(2.9\%) & Relevant & 10.2 & 0.527 &  980  & 0.478  & 5.09 &  3.13  & 19.4   \\
           3094 && Not Relevant &  10.0 & 0.511  & 988  & 0.476  & 5.05 &  3.01  & 18.9   \\
          7907 & & Not Judged &    9.86 & 0.498  & 782  & 0.503  &  4.99 &  3.44  & 18.2   \\

\hline 
 209 & B & & & & & & & & \\
 5 & (2.3\%) & Relevant &  8.40 & 0.360 & 6717  &  0.323   & 4.74   & 32.7  &  21.1   \\
    70 & & Not Relevant & 8.40 & 0.341 & 3933  &  0.346   & 4.82   & 21.5   & 17.7   \\
          134 & & Not Judged &   8.20 & 0.165 & 1481  &  0.335    & 4.62   & 38.9  &  27.1   \\

\hline

13 669 & C & & & & & & & & \\
309 & (2.2\%)   & Relevant &     9.89  & 0.502 & 677  & 0.511  & 5.08  & 7.38 & 20.3  \\
    2516 &      & Not Relevant & 9.87  & 0.482 & 656  & 0.504  & 5.09  & 7.73 & 20.7  \\
   10844  &      & Not Judged &  9.44  & 0.459 & 528  & 0.516  & 5.00  & 10.4 & 20.5  \\

\hline
6006 &  D & & & & & & & & \\
124 & (2.0\%)    & Relevant &     9.63  & 0.480  & 1009  &  0.495  &  4.95 &  5.25 &  18.2  \\
 1278 & & Not Relevant &    9.53  & 0.477  & 1075 &  0.484  &  4.94 &  4.89  & 17.9  \\
 4604 &  & Not Judged &      9.38 &  0.464  & 835 &   0.514 &   4.90  & 6.00 &  17.8  \\
\hline
2613 &  E & & & & & & & & \\
49 & (1.8\%) & Relevant &     10.3  &   0.516  & 1249  & 0.442  & 4.89  & 2.97  & 19.5  \\
      604 &  & Not Relevant &  9.91 &  0.503  & 1228  &  0.446 &  4.90 &  3.06 &  18.9  \\
      1960   & & Not Judged &    9.95 &  0.499  & 855 &  0.486  &  4.86 &  3.24 &   18.7  \\

\hline
3187 & F  & & & & & & & & \\
 48 & (1.5\%) & Relevant &     10.6 &  0.580  & 597  &  0.554  & 5.17 &  4.32  & 21.2  \\
      707 &  &  Not Relevant &  10.1 &  0.484  & 503  & 0.577  & 5.20 &  3.49 &  19.8  \\
      2432 &  &  Not Judged &    9.78  & 0.434  & 367  & 0.600 &  5.12  & 4.53  & 19.6  \\

\hline

21 941 &  G & & & & & & & & \\
183 & (0.8\%) & Relevant &     10.3  & 0.452  & 241 &   0.629 &  5.18  & 6.25  & 20.8  \\
       2526 &  &  Not Relevant &  9.90  & 0.409  & 189  &  0.644 &  5.17  & 7.29  & 20.3  \\
       19232 &  &  Not Judged &    9.55 &  0.388 &  169 &  0.651 &   5.10 &  8.73 &  19.6  \\
\hline
3539 & H  & & & & & & & & \\
21 &  (0.5\%) & Relevant  &     9.24 & 0.397 & 535  & 0.588 & 4.91 &   21.1  & 13.7  \\
       490 &  &  Not Relevant &  8.83  & 0.402 & 643  & 0.543  & 4.85 & 27.0  & 11.7  \\
       3028 &  &  Not Judged &    8.17  & 0.331 & 467  & 0.566 & 4.83 & 27.1  & 14.5  \\
\hline
1096 & I & & & & & & & & \\
  6 & (0.5\%) & Relevant &     9.12  & 0.460 & 377  & 0.603 & 4.35 & 51.1 & 18.7  \\
       145 & &  Not Relevant &  8.33 & 0.275 & 677 & 0.610 & 4.29 & 77.2 & 17.0  \\
       945 &  & Not Judged &    7.12 & 0.150 & 250  & 0.691 & 4.67 & 65.2  & 24.9  \\
\hline
10 925 & J  & & & & & & & & \\
 41 &  (0.3\%) & Relevant &     10.1  & 0.476 & 107 &  0.743 & 5.23 & 6.31 & 22.4 \\
   1052 & &  Not Relevant &  10.4  & 0.330 &  75  & 0.800 & 5.24 & 7.25 & 20.2 \\
  9832 & &  Not Judged &    10.1  & 0.328  & 70 &  0.805 & 5.15 & 8.14 & 19.5 \\
\hline
\end{tabular}
  \end{center}
\caption{Clusters based on stylistic data, and their proportions of relevant documents }\label{Genre}
\end{table}

\section{Defining Genres}

Stylistic variation, as indicated above, is partly an effect of {\em genre}
variation. To get closer to the genres one can expect to find in the corpus
text one issue of Wall Street Journal (910102) was categorized manually
into ten rough categories: articles, business news with, and without
tables, lists of paragraph length items, editorials, letters,
paragraph-length items, ``What's News'' (menu-type lists of one-sentence
items), tables, and single one-sentence items. This was used as a training
set to categorize the entire corpus: simple stylistic measurements for the
hand categorized data -- as shown in the previous section\footnote{With one
extra measure added: digits per character, multiplied by 1000.} -- were
used in a discriminant analysis, and the resulting discriminant functions
were used to automatically categorize the entire corpus. The details of the
method are not important: the result is sloppy in any case. No checking was
made to see how well and consistently the articles were categorized in the
genres given; the idea was simply to have a seed set to cluster the
documents around. In Table~\ref{Genre} some statistics for each category
are shown; the category names have been replaced with letters so as not to
imply the categories are consistent with real life genres.\footnote{The
determined reader will be glad to know that the order of the document
clusters in the table is the same as in the listing of the hand assessed
seed categories in the beginning of the paragraph.}

The hypothesis was that a simple stylistic clustering might well prove
useful thanks to its anchoring in genre, and in spite of this anchoring
being quite tentative. The table shows that there are considerable
differences between the categories in stylistic metrics -- unsurprisingly,
since they have been clustered to maximise that difference -- but more
importantly, the categories show considerable differences in how large a
proportion of the documents are relevant, and most importantly, in {\em
how} the relevant documents differ from the nonrelevant ones
stylistically. For instance, whereas in category A, relevant documents will
have longer sentences on average than non-relevant and non-retrieved documents,
in categories C and H the relevant documents will have shorter sentences;
and whereas most categories prefer documents with a low type-token ratio,
category H prefers documents with a high ratio.\footnote{These differences
are significant on a better than 95\% level, by Mann Whitney U; most
numbers in Table~\ref{Genre} have not been checked, however.}

\section{Conclusions}

Texts differ in style. In this experiment, automatically retrieved texts
differed from non-retrieved texts along several simple stylistic metrics.
This shows that either 1) retrieval mechanisms are biased for style, or
more likely, 2) style and topic go hand in hand. Neither of these results
are surprising. Nonetheless, they may be a useful point to note for
information retrieval system designers.

What is more interesting, and a good starting point for user-oriented
information retrieval studies is utilizing this type of measure in
distinguishing {\em interesting} texts from {\em less interesting} ones. This
will entail analyzing the tasks and expectations of users; this experiment
shows that for a certain set of users and for a certain scenario a clear
bias towards a certain types or genres of text can be found, namely
the one between relevant and non-relevant in the experiment.

The experiment also shows that stylistically determined genres or
functional styles are different as regards potential usefulness for the
queries tested, and that the distinctions between relevant and non-relevant
differ between genres. 

The differences between relevant and non-relevant texts found should not be
taken as general results: while useful in a TREC context, as shown by the
results from Cornell, they are clearly an effect of the task, corpus, and
assessors. These results should be taken as a starting point in
investigating how situations affect measures of stylistic variation.

\section*{Acknowledgments}

This paper reflects work done for the General Electric, New York
University, Lockheed Martin, and Rutgers University project for the fifth
Text REtrieval Conference. The experiments have benefited greatly from
discussions with my project colleagues: Louise Guthrie, Fang Lin, Jos\'e
P\'erez-Carballo, and Tomek Strzalkowski. Without the suggestions for
improvements and corrections -- some of which I heeded -- given me by Ralph
Grishman and Slava Katz at New York University, these experiments would not
hold up at all. Thank you all!

\end{document}